\documentclass[runningheads,a4paper]{llncs}

\usepackage{graphicx}
\usepackage{epstopdf}
\usepackage{caption}
\usepackage{subcaption}

\usepackage[overlay,absolute]{textpos}

\begin{document}

\begin{textblock*}{10in}(38mm, 10mm)
{\textbf{Ref:} \emph{International Conference on Artificial Neural Networks (ICANN)}, Springer LNCS, }
\end{textblock*}
\begin{textblock*}{10in}(38mm, 15mm)
{Vol.~9887, pp.~88--96, Barcelona, Spain, 2016.}
\end{textblock*}

\begin{textblock*}{10in}(64mm, 26mm)
{\fbox{\large Winner of Best Paper Award in ICANN 2016}}
\end{textblock*}

\author{}
\institute{}
\author{Eli (Omid) David\inst{1,2} \and Nathan S. Netanyahu\inst{2,3} \and Lior Wolf\inst{1}}

\institute{
	The Blavatnik School of Computer Science,
	Tel Aviv University, Tel Aviv, Israel\\
	\email{mail@elidavid.com, wolf@cs.tau.ac.il}\\
	\and
	Department of Computer Science,
	Bar-Ilan University, Ramat-Gan, Israel\\
	\email{nathan@cs.biu.ac.il}\\
	\and
	Center for Automation Research,
	University of Maryland,
	College Park, MD, USA\\
	\email{nathan@cfar.umd.edu} 
	}

\title{DeepChess: End-to-End Deep Neural Network \\for Automatic Learning in Chess}

\titlerunning{DeepChess: End-to-End Deep Neural Network for Automatic Learning in Chess}
\authorrunning{E.O.~David et al.}

\maketitle 
\vspace*{-10pt} 

\begin{abstract}

We present an end-to-end learning method for chess, relying on deep neural networks. Without any a priori knowledge, in particular without any knowledge regarding the rules of chess, a deep neural network is trained using a combination of unsupervised pretraining and supervised training. The unsupervised training extracts high level features from a given position, and the supervised training learns to compare two chess positions and select the more favorable one. The training relies entirely on datasets of several million chess games, and no further domain specific knowledge is incorporated.

The experiments show that the resulting neural network (referred to as DeepChess) is on a par with state-of-the-art chess playing programs, which have been developed through many years of manual feature selection and tuning. DeepChess is the first end-to-end machine learning-based method that results in a grandmaster-level chess playing performance.

\end{abstract}

\section{Introduction}
\vspace*{-6pt} 
Top computer chess programs are based typically on manual feature selection and tuning of their evaluation function, usually through years of trial and error. While computer chess is one of the most researched fields within AI, machine learning has not been successful yet at producing grandmaster level players.

In this paper, we employ deep neural networks to learn an evaluation function \emph{from scratch}, without incorporating the rules of the game and using no manually extracted features at all. Instead, the system is trained from end to end on a large dataset of chess positions.

Training is done in multiple phases. First, we use deep unsupervised neural networks for pretraining. We then train a supervised network to select a preferable position out of two input positions. This second network is incorporated into a new form of alpha-beta search.  A third training phase is used to compress the network in order to allow rapid computation. 

Our method obtains a grandmaster-level chess playing performance, on a par with top state-of-the-art chess programs. To the best of our knowledge, this is the first machine learning-based method that is capable of learning from scratch and obtains a grandmaster-level performance.

\section{\label{sec:learning} Previous Work}
\vspace*{-6pt} 

Chess-playing programs have been improved significantly over the past several decades. While the first chess programs could not pose a challenge to even a novice player, the current advanced chess programs have been outperforming the strongest human players, as the recent man vs.~machine matches clearly indicate. Despite these achievements, a glaring deficiency of today's top chess programs is their severe lack of a learning capability (except in most negligible ways, e.g., ``learning'' not to play an opening that resulted in a loss, etc.). 

During more than fifty years of research in the area of computer games, many learning methods have been employed in several games. \emph{Reinforcement learning} has been successfully applied in backgammon \cite{tesauro92} and checkers \cite{schaeffer01}. Although reinforcement learning has also been applied to chess \cite{baxter00,lai15}, the resulting programs exhibit a playing strength at a human master level at best, which is substantially lower than the grandmaster-level state-of-the-art chess programs. These experimental results confirm Wiering's \cite{wiering95} formal arguments for the failure of reinforcement learning in rather complex games such as chess. Very recently, a combination of a \emph{Monte-Carlo search} and deep learning resulted in a huge improvement in the game of Go \cite{silver16}. However, Monte-Carlo search is not applicable to chess, since it is much more tactical than Go,  e.g., in a certain position, all but one of the moves by the opponent may result in a favorable result, but one refutation is sufficient to render the position unfavorable.

In our previous works, we demonstrated how genetic algorithms (GA's) could be applied successfully to the problem of automatic evaluation function tuning when the features are initialized randomly \cite{david08c,david09a,david11,david14}. Although to the best of our knowledge, these works are the only successful automatic learning methods to have resulted in grandmaster-level performance in computer chess, they do not involve learning the features themselves from scratch. Rather, they rely on the existence of a manually created evaluation function, which consists already of all the required features (e.g., queen value, rook value, king safety, pawn structure evaluation, and many other hand crafted features). Thus, GAs are used in this context for \emph{optimization} of the weights of existing features, rather than for \emph{feature learning} from scratch.

\section{\label{sec:deepchess} Learning to Compare Positions}
\vspace*{-6pt} 

The evaluation function is the most important component of a chess program. It receives a chess position as an input, and provides a score as an output. This score represents how good the given position is (typically from White's perspective). For example, a drawish position would have a score close to $0$, a position in which white has two pawns more than black would have a score of $+2$, and a position in which black has a rook more than white, would be scored around $-5$. A good evaluation function considers typically a large number (i.e., on the order of hundreds and even thousands) of properties in addition to various piece-related parameters, such as king safety, passed pawns, doubled pawns, piece centrality, etc. The resulting score is a linear combination of all the selected features. The more accurately these features and their associated values capture the inherent properties of the position, the stronger the corresponding chess program becomes.

In this paper, we are interested in developing such an evaluation function from scratch, i.e., with absolutely no a priori knowledge. As a result, we do not provide our evaluation function with any features, including any knowledge about the rules of chess. Thus, for our training purposes, we are limited to observing databases of chess games with access only to the results of the games (i.e., either a win for White or Black, or a draw).

Since the real objective of an evaluation function is to perform relative comparisons between positions, we propose a novel training method around this concept. The model receives two positions as input and learns to predict which position is better. During training, the input pair is selected as follows: One position is selected at random from a game which White eventually won and the other from a game which Black eventually won. This relies on the safe assumption that, on average, positions taken from games that White won are preferable (from White's perspective) to those taken from games that White lost. Additionally, the proposed approach allows for the creation of a considerably larger training dataset. For example, if we have a million positions from games that White had won, and a million positions from games that White had lost, we can create $2 \times 10^{12}$ training pairs (multiplied by 2 because each pair can be used twice, as [win, ~loss] and [loss, ~win]).

\begin{figure}[t]
	\centering
	\includegraphics[height=2.9in, width=5in]{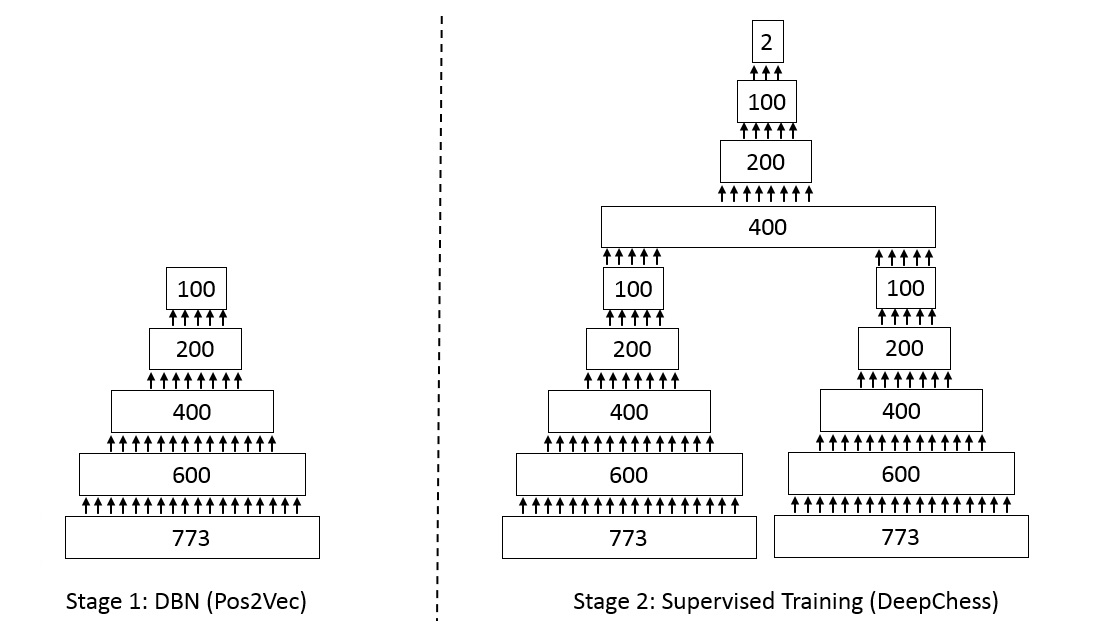}
	\caption{Architecture illustration of DeepChess.}
	\label{fig:deepchess}
\end{figure}

Our approach consists of multiple stages. First, we train a deep autoencoder on a dataset of several million chess positions. This deep autoencoder functions as a nonlinear feature extractor. We refer to this component as \emph{Pos2Vec}, since it converts a given chess position into a vector of values which represent the high level features. In the second phase, we use two copies of this pretrained Pos2Vec side by side, and add fully connected layers on top of them, with a 2-value softmax output layer. We refer to this structure as \emph{DeepChess}. It is trained to predict which of the two positions results in a win. Note that similar to the most successful object detection methods \cite{fasterrcnn}, we found a 2-value output to outperform one binary output. Figure \ref{fig:deepchess} illustrates the neural network architecture.

\medskip
\noindent{\bf Dataset:}
We employed the games dataset of CCRL (\url{www.computerchess.org.uk/ccrl}), which contains 640,000 chess games, out of which White won 221,695 games and Black won 164,387 games, the remaining games ended in a draw. Our experiments show that the inclusion of games that ended in a draw is not beneficial, so we only use games which ended in a win.

From each game we randomly extracted ten positions, with the restriction that the selected position cannot be from one of the first five moves in the game, and that the actual move played in the selected position is not a capture. Capture moves are misleading as they mostly result in a transient advantage since the other side is likely to capture back right away. The dataset thus contains 2,216,950 positions from games which White won ($W$ positions), and 1,643,870 positions from games which White lost ($L$ positions), for a total of 3,860,820 positions.

Each position is converted to a binary bit-string of size 773. There are two sides (White and Black), 6 piece types (pawn, knight, bishop, rook, queen, and king), and 64 squares. Therefore, in order to represent a position as a binary bit-string, we would require $2 \times 6 \times 64 = 768$ bits (this is known as \emph{bitboard} representation).  There are an additional five bits that represent the side to move (1 for White and 0 for Black) and castling rights (White can castle kingside, White can castle queenside, Black can castle kingside, and Black can castle queenside).

\medskip
\noindent{\bf Training Pos2Vec:} 
We first trained a deep belief network (DBN) \cite{bengio07b}, which would later serve as the initial weights for supervised training. The DBN is based on stacked autoencoders which are trained using layer-wise unsupervised training. The network consists of five fully connected layers of sizes: 773--600--400--200--100. We initially trained the first layer (i.e., a 3-layer (773--600--773) autoencoder), before fixing its weights and training the weights of a new (600--400--600) autoencoder, and so on.

We used a random subset of 2,000,000 chess positions for training the DBN, of which 1,000,000 were White win ($W$) positions and 1,000,000 were Black win ($L$) positions. The DBN uses a rectified linear unit (ReLU), i.e., $f(x)=max(0,x)$, and a learning rate that starts from 0.005 and is multiplied by 0.98 at the end of each epoch. No regularization is used. The DBN is trained for 200 epochs.

\medskip
\noindent{\bf Training DeepChess:}
As described earlier, this Siamese network is the core component of our method. We used the previously trained Pos2Vec DBN as the initial weights for the supervised network. Placing two disjoint copies of Pos2Vec side by side, we added on top of them four fully connected layers of size 400, 200, 100, and 2, which are connected to both Pos2Vec components. The first five layers of Pos2Vec thus serve as high level feature extractors, and the last four layers compare the features of the positions to determine which one is better. 

During the supervised training phase, the entire network including the Pos2Vec parts is modified. We tie the weights of the two Pos2Vec-based feature extraction components, i.e., we use shared weights. 

We trained this network for 1000 epochs. In each epoch, we created 1,000,000 random input pairs, where each pair consists of one position selected at random from the 2,116,950 $W$ positions, and one position selected at random from the 1,543,870 $L$ positions. (we set aside 100,000 $W$ positions and 100,000 $L$ positions for validation). The pair is then randomly ordered as either ($W,L$) or ($L,W$). Since the number of potential training pairs is $6.5 \times 10^{12}$, virtually all training samples in each epoch are new, thus guaranteeing that no overfitting would take place. For this reason, we do not use any regularization term. The activation used in all layers is the ReLU function. The learning rate starts from 0.01, and is multiplied by 0.99 after each epoch. The cross entropy loss is used. 
The training and validation accuracies obtained were 98.2\% and 98.0\%, respectively. This is remarkable, considering that no a priori knowledge of chess, including the very rules of the games are provided. 

\medskip
\noindent{\bf Improving Inference Speed by Network Distillation:} 
Before incorporating the trained network into a chess program and evaluating its performance, we first had to address the problem that the network is too computationally expensive in prediction (inference) mode, running markedly slower than a typical evaluation function in a chess program. Several previous works have demonstrated how a considerably smaller neural network could be trained to mimic the behavior of a much more complex neural network \cite{hinton14,romero15}. These network compression or distilling approaches train the smaller network to produce the same output as the larger network (learning from soft targets).

We first trained a smaller four-layer network of 773--100--100--100 neurons to mimic the feature extraction part of DeepChess, which consists of the five layers 773--600--400--200--100. We then added three layers of 100--100--2 neurons (originally 400--200--100--2) and trained the entire network to mimic the entire DeepChess network..

Further optimization was achieved by realizing that while most of the weights are concentrated in the first layer of the two Pos2Vec components (733--100 layer), there are at most 32 chess pieces in a given position and less than 5\% of the weights in the input layer would be activated. Thus the amount of floating point operations required to be performed during inference is much reduced.

Table~\ref{tab:validation} summarizes the validation results post compression. 
The distilled network is comparable to the full original network. When training from scratch using the smaller network size (with pretraining but without first training the larger network and then distilling it), the performance is much reduced.

\section{A Comparison-Based Alpha-Beta Search}
\vspace*{-6pt} 

Chess engines typically use the alpha-beta search algorithm \cite{knuth75}. Alpha-beta is a depth-first search method that prunes unpromising branches of the search tree earlier, improving the search efficiency. A given position is the root of the search tree, and the legal moves for each side create the next layer nodes. The more time available, the deeper this search tree can be processed, which would result in a better overall playing strength. At leaf nodes, an evaluation function is applied.

In an alpha-beta search, two values are stored; $\alpha$ which represents the value of the current best option for the side to move, and $\beta$ which is the negative $\alpha$ of the other side. For each new position encountered if $value > \alpha$, this value would become the new $\alpha$, but if $value > \beta$, the search is stopped and the search tree is pruned, because $value > \beta$ means that the opponent would not have allowed the current position to be reached (better options are available, since  $value > \beta$ is equivalent to $-value < \alpha$ for the other side). Given a branching factor of $B$ and search depth $D$, alpha-beta reduces the search complexity from $B^D$ for basic DFS, to $B^{D/2}$.

In order to incorporate DeepChess, we use a novel version of an alpha-beta algorithm that does not require any position scores for performing the search. Instead of $\alpha$ and $\beta$ values, we store positions $\alpha_{pos}$ and $\beta_{pos}$. For each new position, we compare it with the existing  $\alpha_{pos}$ and $\beta_{pos}$ positions using DeepChess, and if the comparison shows that the new position is better than $\alpha_{pos}$, it would become the new $\alpha_{pos}$, and if the new position is better than $\beta_{pos}$, the current node is pruned. Note that since DeepChess always compares the positions from White's perspective, when using it from Black's perspective, the predictions should be reversed.

\medskip
\noindent{\bf Position hashing:} When searching a tree of possible moves and positions, many of the positions appear repeatedly in different parts of the search tree, since the same position can arise in different move orders. To reduce the required computation, we store a large hash table for positions and their corresponding feature extraction values. For each new position, we first query the hash table, and if the position has already been processed, we reuse the cached values. Since we use a symmetric feature extraction scheme, where the weights are shared, each position needs only be stored once.

\section{Experiments}
\vspace*{-6pt} 

We provide both quantitative and qualitative results. 
\vspace*{-6pt} 

\subsection{Static Understanding of Chess Positions}
\vspace*{-3pt} 

In order to measure the chess understanding of DeepChess, we ran it on a manually generated dataset consisting of carefully designed inputs. Each input pair in this dataset contains two nearly identical positions, where one contains a certain feature and the other one does not. Starting from simple piece values (e.g., two identical positions where a piece is missing from one), to more complex imbalances (e.g., rook vs. knight and a bishop), the predictions of DeepChess show that it has easily learned all of the basic concepts regarding piece values. We then measured more subtle positional features, e.g., king safety, bishop pair, piece mobility, passed pawns, isolated pawns, doubled pawns, castling rights, etc. All of these features are also well understood by DeepChess. 

\begin{figure}
\centering
         \begin{subfigure}[t]{0.23\textwidth}
                \centering
                \includegraphics[width=\textwidth]{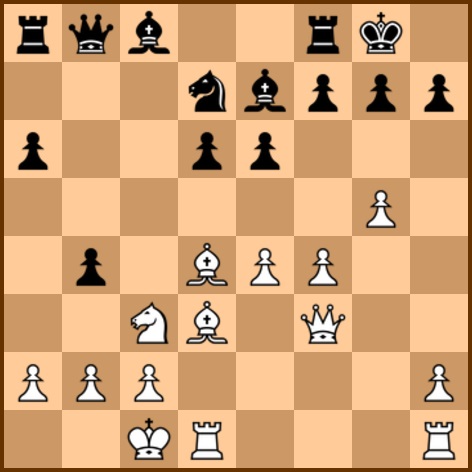}
                \caption*{Tal - Larsen\\Move: \textbf{Nd5}}
        \end{subfigure}
        ~
        \begin{subfigure}[t]{0.23\textwidth}
                \centering
                \includegraphics[width=\textwidth]{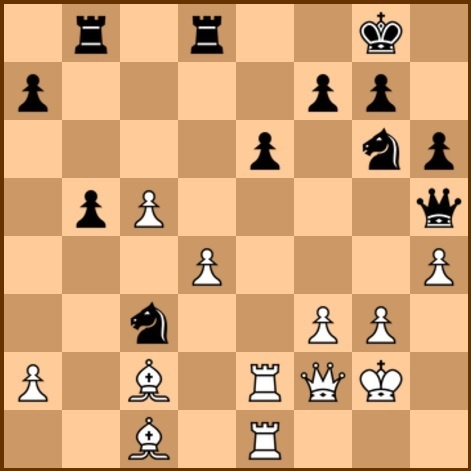}
                \caption*{Aronian - Leko\\Move: \textbf{Re5}}
        \end{subfigure}
        ~
        \begin{subfigure}[t]{0.23\textwidth}
                \centering
                \includegraphics[width=\textwidth]{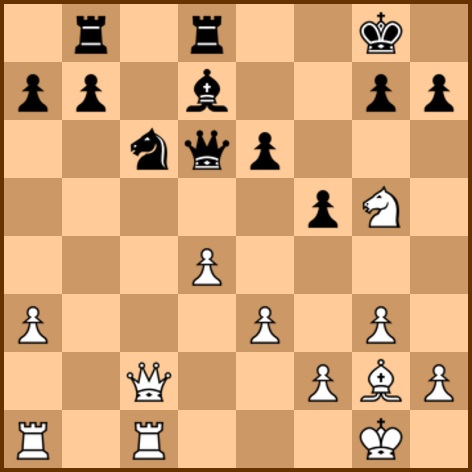}
                \caption*{Alekhine -Golombek\\Move: \textbf{d5}}
        \end{subfigure}
        ~
        \begin{subfigure}[t]{0.23\textwidth}
                \centering
                \includegraphics[width=\textwidth]{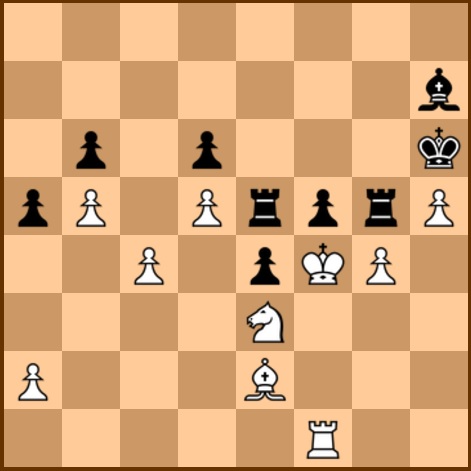}
                \caption*{Seirawan - Kozul\\Move: \textbf{c5}}
        \end{subfigure}
\vspace*{-6pt} 
        \caption{Examples where DeepChess prefers to play the same positional sacrifices that were played by grandmasters. It is White's turn to move in all the above positions.}
\vspace*{-10pt} 
        \label{fig:positions}
\end{figure}

More interestingly, DeepChess has learned to prefer positions with dynamic attacking opportunities even when it has less material. In many cases, it prefers a position with one or two fewer pawns, but one that offers non-material positional advantages.  This property has been associated with human grandmasters, and has always been considered an area in which computer chess programs were lacking. While the scores of current evaluation functions in state-of-the-art chess programs are based on a linear combination of all the features present, DeepChess is a non-linear evaluator, and thus has a far higher potential for profound understanding of chess positions (also similar to human grandmaster analysis of positions). Figure~\ref{fig:positions} shows a few examples where this preference of DeepChess for non-materialistic advantages leads to favoring positional sacrifices, as played by human grandmasters.

\subsection{Playing Strength vs. State-of-the-Art Competitors}
\vspace*{-3pt} 

We used the \textsc{Falcon} chess engine as a baseline for our experiments. \textsc{Falcon} is a grandmaster-level chess program, which has successfully participated in several World Computer Chess Championships (WCCCs); in particular, it won second place at the World Computer Speed Chess Championship in 2008. \textsc{Falcon}'s extensive evaluation function consists of more than 100 parameters, and its implementation contains several thousands of lines of code. 

Despite all the computational improvements mentioned earlier for DeepChess, and numerous other implementation improvements which result in substantial additional computational speedup, DeepChess is still four times slower than \textsc{Falcon}'s own evaluation function. Nevertheless, we incorporate DeepChess into \textsc{Falcon}, completely replacing the evaluation function of the program. 

To measure the performance of DeepChess, we conducted a series of matches against \textsc{Falcon}, and also against the chess program \textsc{Crafty}. \textsc{Crafty} has successfully participated in numerous WCCCs, and is a direct descendant of Cray Blitz, the WCCC winner of 1983 and 1986. It has been frequently used in the literature as a standard reference. 

Each of the matches of DeepChess vs.~\textsc{Falcon} and \textsc{Crafty} consisted of 100 games under a time control of 30 minutes per game for each side.  Table~\ref{tab:search-matches} provides the results. As can be seen, DeepChess is on a par with \textsc{Falcon}. \textsc{Falcon} uses a manually tuned evaluation function developed over nearly ten years, containing more than a hundred parameters which grasp many subtle chess features. And yet, without any chess knowledge whatsoever (not even basic knowledge as the rules of chess), our DeepChess method managed to reach a level which is on a par with the manually tuned evaluation function of \textsc{Falcon}. The results also show that DeepChess is over 60 Elo~\cite{elo78} stronger than \textsc{Crafty}, a program which has won two WCCCs and has been manually tuned for thirty years.

\begin{table}[t]
\parbox{.38\linewidth}{
\centering
\setlength{\tabcolsep}{1.1em} 
	{\renewcommand{\arraystretch}{1.2}
    \begin{tabular}{|l|c|}
    \hline
    Method&Accuracy\\
    \hline
    \hline
    Uncompressed & 98.0\% \\
    Compressed & 97.1\%\\
    Small  & 95.4\% \\
    \hline
    \end{tabular}
    }
    \medskip
    \caption{\label{tab:validation} Validation accuracy of the Uncompressed and compressed networks, and a small network trained from scratch.}
    }
\hfill
\parbox{.57\linewidth}{
\centering
\setlength{\tabcolsep}{0.4em} 
		{\renewcommand{\arraystretch}{1.2}
		\begin{tabular}{|l|c|c|}
			\hline
			Match & Result & RD\\
			\hline
            \hline
			DeepChess 30min - \textsc{Crafty} & 59.0 - 41.0 & $+63.2$\\
DeepChess 30min - \textsc{Falcon} & 51.5 - 48.5 & $+10.4$\\
            DeepChess 120min - \textsc{Falcon} & 63.5 - 36.5 & $+96.2$\\
            \hline
		\end{tabular}
        }
        \medskip
\caption{	\label{tab:search-matches} DeepChess vs.~\textsc{Falcon} and \textsc{Crafty} (RD is the Elo rating difference). Time control: 30 minutes per game for \textsc{Falcon} and \textsc{Crafty}. 30 minutes or two hours for DeepChess.}
}
\end{table}

DeepChess performs on a par with \textsc{Falcon} despite the fact that it is four times slower. We ran a separate experiment where we allowed DeepChess to use four times more time than \textsc{Falcon} (2 hours vs 30 minutes). Running 100 such matches, DeepChess resoundingly defeated \textsc{Falcon} with a result of 63.5 - 36.5, corresponding to a 96 Elo performance difference. This shows that DeepChess is actually not on par with \textsc{Falcon}'s evaluation function, but is considerably superior to it. In order to utilize the full potential of this enhanced chess understanding, it is critical to decrease the runtime of the neural network in the inference mode.

\section{\label{sec:conclusions}Concluding Remarks}
\vspace*{-6pt} 
We presented the first successful end-to-end application of machine learning in computer chess. Similarly to human chess masters, DeepChess does not assign numeral evaluation values to different positions, but rather, \emph{compares} different positions that may arise, and opts for the most promising continuation. 

Having observed the playing style of DeepChess, we note that it plays very aggressively, often sacrificing pieces for long term positional gains (i.e., non-tactical gains). This playing style resembles very much the playing style of human grandmasters. While computer chess programs have long been criticized for being materialistic, DeepChess demonstrates the very opposite by exhibiting an adventurous playing style with frequent positional sacrifices.

\end{document}